\documentclass[runningheads]{llncs}
\usepackage{graphicx}
\usepackage{subfigure}
\usepackage{graphicx}
\usepackage{float}
\usepackage{multirow}
\usepackage{placeins}
\usepackage{amssymb}
\usepackage{amsmath}

\begin{document}
\title{Region Proposal Rectification Towards Robust Instance Segmentation of Biological Images}
%

\author{
Qilong Zhangli\inst{1},
Jingru Yi\inst{1}, 
Di Liu\inst{1},
Xiaoxiao He\inst{1}, 
Zhaoyang Xia\inst{1}, 
Qi Chang\inst{1}, 
Ligong Han\inst{1}, 
Yunhe Gao\inst{1},
Song Wen\inst{1}, 
Haiming Tang\inst{2}, 
He Wang\inst{2}, 
Mu Zhou\inst{3}, 
Dimitris Metaxas\inst{1} 
}

\authorrunning{Q. Zhangli et al.}

\institute{Department of Computer Science, Rutgers University, NJ, USA. \and School of Medicine, Yale University, CT, USA.\and SenseBrain Research, CA, USA.}

\maketitle              
\begin{abstract}
Top-down instance segmentation framework has shown its superiority in object detection compared to the bottom-up framework. While it is efficient in addressing over-segmentation, top-down instance segmentation suffers from over-crop problem. However, a complete segmentation mask is crucial for biological image analysis as it delivers important morphological properties such as shapes and volumes. In this paper, we propose a region proposal rectification (RPR) module to address this challenging incomplete segmentation problem. In particular, we offer a progressive ROIAlign module to introduce neighbor information into a series of ROIs gradually. The ROI features are fed into an attentive feed-forward network (FFN) for proposal box regression. With additional neighbor information, the proposed RPR module shows significant improvement in correction of region proposal locations and thereby exhibits favorable instance segmentation performances on three biological image datasets compared to state-of-the-art baseline methods. Experimental results demonstrate that the proposed RPR module is effective in both anchor-based and anchor-free top-down instance segmentation approaches, suggesting the proposed method can be applied to general top-down instance segmentation of biological images.

\end{abstract}
\section{Introduction}
Instance segmentation of biological images is challenging due to the uneven texture, unclear boundary, and touching problem of the biological objects, as widely seen in plant phenotyping \cite{minervini2016finely} and computational pathology \cite{irshad2014crowdsourcing,liu2020dispersion,liu2019dispersion,pantanowitz2010digital,chang2022deeprecon}. Current research efforts mainly rely on two strategies including bottom-up~\cite{chen2017dcan,oda2018besnet} and top-down~\cite{he2017mask,lee2020centermask,yi2019multi,yi2019attentive,hu2020harnessing,sayadi2022harnessing} frameworks. Bottom-up approach performs semantic segmentation first on the whole input images. Instance masks are obtained subsequently according to features such as contours \cite{chen2017dcan,oda2018besnet,zhang2021multi}, morphological shapes \cite{schmidt2018cell,zhou2019cia,liu2021label}, and pixel similarities \cite{payer2018instance,liu2021refined,he2019effective,gao2021utnet,gao2022multi,liu2022transfusion}. This framework suffers from careful feature designs to avoid over-/under-segmentation. Meanwhile, top-down framework utilizes global objectness features and generates region proposals as a crucial initialization step \cite{he2017mask,lee2020centermask,yi2019attentive,yi2020object}. A region proposal is represented as a bounding box around each object. Semantic segmentation is performed subsequently on region of interest (ROI) features inside each region proposal. The ROI features are usually cropped through ROIAlign \cite{he2017mask,lee2020centermask}.

\begin{figure}[thb!]
\centering
\subfigure[Original]{
    \label{Fig1.sub.1}
    \includegraphics[width=0.2\linewidth]{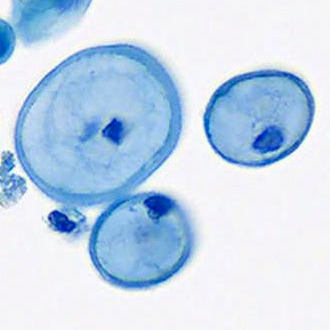}}
\subfigure[Ground Truth]{
    \label{Fig1.sub.2}
    \includegraphics[width=0.21\linewidth]{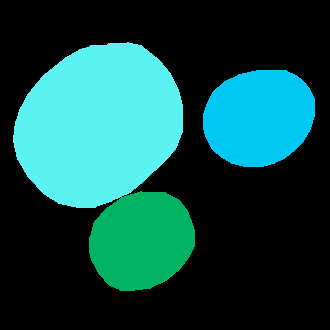}}
\subfigure[Baseline]{
    \label{Fig1.sub.3}
    \includegraphics[width=0.21\linewidth]{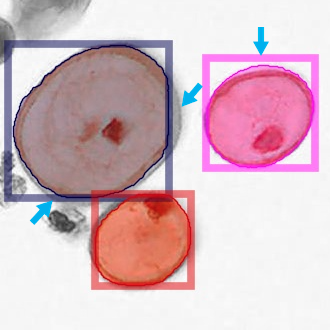}}
\subfigure[Baseline+RPR]{
    \label{Fig1.sub.4}
    \includegraphics[width=0.21\linewidth]{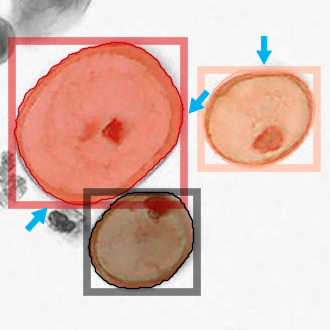}}
\caption{Illustration of incomplete segmentation problem for top-down instance segmentation on urothelial cell images. (a) and (b) are the original image and ground truth annotation. In (c), a top-down baseline method (i.e., CenterMask \cite{lee2020centermask}) outputs incomplete segmentation masks due to inaccurate region proposals (i.e., bounding boxes). In (d), with the proposed region proposal rectification (RPR) module, the box boundaries are corrected and the segmentation masks are intact.
}
\label{Figure1}
\end{figure}

Top-down framework highlights its superiority in object detection while it brings significant challenges. First, an ROI region could contain noisy pixels from neighbor objects, which is difficult to be suppressed by network. 
Second, a region proposal restricts the ROI space for intact and high-quality segmentation. A slight variation in region proposal location can cut off object boundary and result in incomplete mask (see Fig.~\ref{Figure1}). As a segmentation mask 
carries object's morphological properties (e.g., shape, volume) for biological image analysis, generating a complete 
segmentation mask is of great 
importance.

In this paper, to address the challenging incomplete segmentation problem in top-down instance segmentation, we propose a region proposal rectification (RPR) module which involves two components: a progressive ROIAlign and an attentive feed-forward network (FFN). Progressive ROIAlign
incorporates neighbor information into a series of ROI features and attentive FFN studies spatial relationships of features and delivers rectified region proposals. Segmentation masks are generated subsequently with the rectified proposals. On three biological image datasets, RPR shows significant improvement in region proposal location rectification and achieves favorable performances in instance segmentation for both anchor-based and anchor-free approaches (e.g., Mask R-CNN \cite{he2017mask} and CenterMask \cite{lee2020centermask}). Results suggest that the proposed method is able to serve for general top-down instance segmentation of biological images. To the best of our knowledge, we are the first work that adopts this Transformer-based architecture on the refinement of region proposals in biological instance segmentation. 

\begin{figure} [thb!]
\begin{center}
\includegraphics[width=0.9\linewidth]{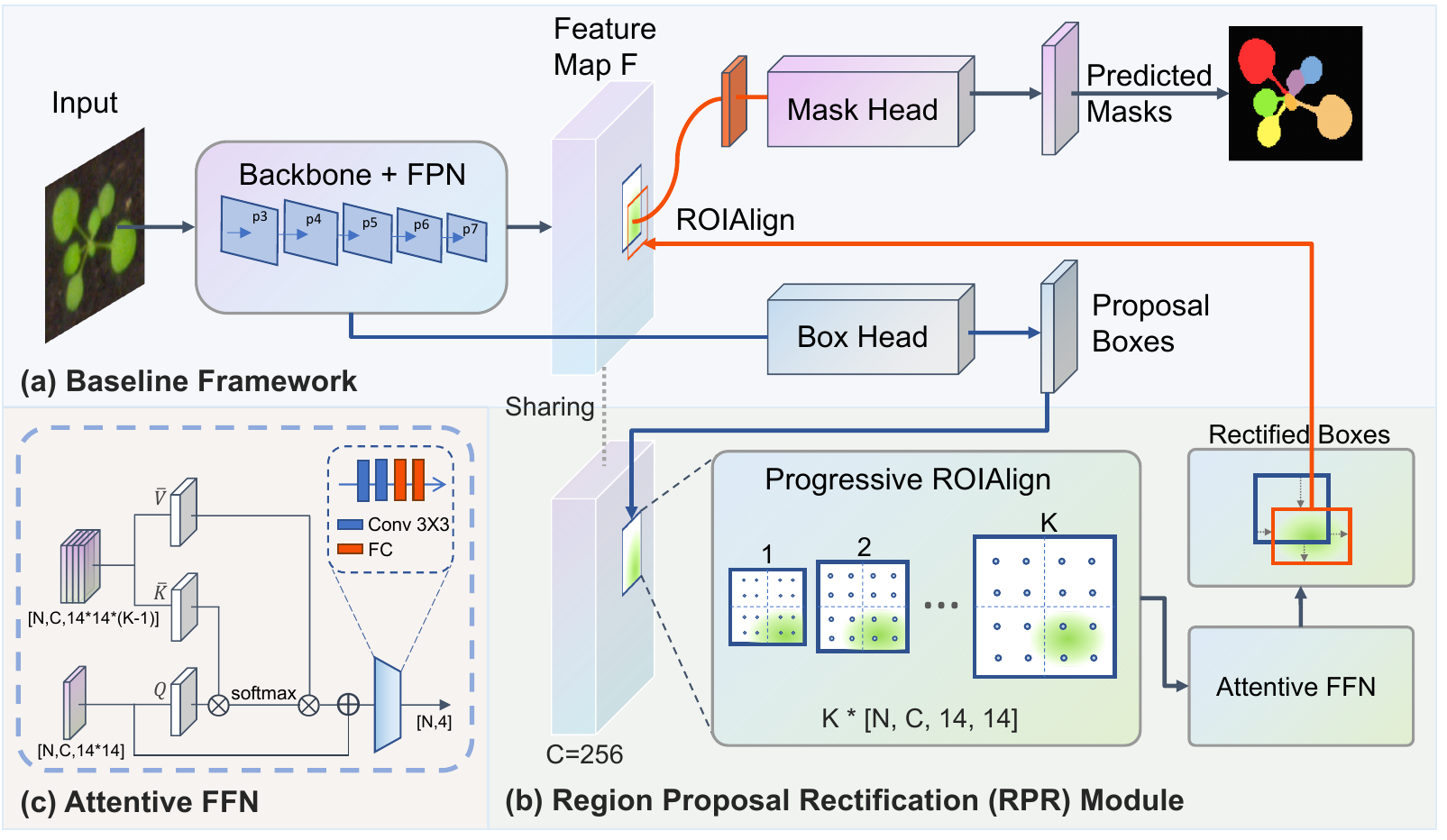}
\end{center}
   \caption{Diagram of top-down instance segmentation method with (a) a representative baseline framework; (b) region proposal rectification (RPR) module; (c) attentive FFN. Input image resolution is resized to 800$\times$800. We use ResNet50 \cite{he2016deep} as backbone network. Box head generates region proposal boxes based on output features from backbone+FPN. 
   For each region proposal box, progressive ROIAlign module extracts a series of ROIs and feeds them to attentive FFN for box rectification. 
   Mask head makes segmentation on ROI features cropped by ROIAlign based on rectified boxes.
   }
\label{Figure2}
\end{figure}

\section{Method}
The overview of the proposed approach is shown in Fig.~\ref{Figure2}, which contains three main components: the baseline framework (Fig.~\ref{Figure2}a), the region proposal rectification (RPR) module (Fig.~\ref{Figure2}b), and the attentive feed-forward network (FFN) (Fig.~\ref{Figure2}c). 
Component details are illustrated in the following sections.

\subsection{Baseline Framework}
Fig.~\ref{Figure2}a shows a representative top-down instance segmentation framework that involves three steps: (1) feature extraction; (2) region proposal generation; and (3) mask prediction. Feature extraction employs a backbone network (i.e. deep neural network) to extract multi-scale features from input image. Feature pyramid network (FPN) \cite{lin2017feature} is usually adopted to build high-level semantic features at all scales. Based on the learned feature maps $F$, region proposal boxes are generated through a box head (see Fig.~\ref{Figure2}a). With proposal boxes, ROIAlign \cite{he2017mask} extracts ROIs from $F$ and a mask head performs segmentation on ROIs subsequently. In this paper, we adopt two representative top-down instance segmentation methods as our baseline: Mask R-CNN \cite{he2017mask} and CenterMask \cite{lee2020centermask}. Mask R-CNN employs anchor-based region proposal network (RPN) \cite{ren2015faster} 
while CenterMask uses anchor-free FCOS \cite{tian2019fcos} to provide proposal boxes. We use a backbone (e.g., ResNet50 \cite{he2016deep}) with FPN for feature extraction. The box head and mask head share the same structures as Mask R-CNN and CenterMask.

\subsection{Region Proposal Rectification}
As mentioned above, since the baseline approaches make segmentation on extracted ROIs, it suffers from an incomplete segmentation problem (see Fig.~\ref{Figure1}c). To address this problem, in this paper, we propose a region proposal rectification (RPR) module. The flowchart of the RPR module is shown in Fig.~\ref{Figure2}b. The RPR module mainly contains two sub-modules: a progressive ROIAlign module and an attentive FFN module.

\subsubsection{Progressive ROIAlign} ROIAlign \cite{he2017mask} is proposed in Mask R-CNN to extract well-aligned ROI features based on proposal boxes. Although it improves the misalignment problem in ROI Pooling \cite{girshick2015fast}, the cropped ROIs suffers from incomplete segmentation problem (see Fig.~\ref{Figure1}c). 
To relieve this problem, Mask R-CNN \cite{he2017mask} attempts a second-stage box regression based on cropped ROI features. However, the limited view of ROIs lacks global knowledge and therefore brings difficulty for the network to correct proposal boxes. In this study, we propose to rectify the region proposal locations from features with an expanded view. In particular, a progressive ROIAlign module is employed to expand proposal regions progressively and a batch of ROIs features are obtained for the following proposal rectification. We show that by introducing neighbor features, the network gains more related knowledge and is able to rectify the proposal locations.

As shown in Fig.~\ref{Figure2}b, progressive ROIAlign extracts ROIs features (ROIs$^{\text{rec}}=\{$ROI$^{\text{rec}}_i\in\mathbb{R}^{N\times  C\times14\times14}\}_{i=1}^{K}$) from proposal boxes $\mathcal{B}\in\mathbb{R}^{N\times 4}$, where $C$ indicate feature channels, $K$ represents the number of extracted ROIs from each proposal box, $N$ is the total number of proposal boxes. Given input feature map $F$ and proposal boxes $\mathcal{B}$, the extracted ROI features can be expressed as:
\begin{align}
    \text{ROIs}^{\text{rec}} &= \text{progressive ROIAlign}(F, \mathcal{B}).
\end{align}
For each proposal box $B=(x_1,y_1,x_2,y_2) \in \mathcal{B}$, where $(x_1,y_1)$ and $(x_2,y_2)$ represent the top-left and bottom-right corner points of $B$, the progressive ROIAlign expands $B$ in $K$ iterations with a dilation rate $\rho$ and generates expanded proposal boxes $B^{\text{expand}}=\{B^{\text{expand}}_i\}_{i=1}^K$. Each $B^{\text{expand}}_i$=($\hat{x}_{i_1}$, $\hat{y}_{i_1}$, $\hat{x}_{i_2}$, $\hat{y}_{i_2})$ is formulated as:
\begin{align}
\begin{split}
    \hat{x}_{i_1} = \max(x_1\times (1-r_i),~0),& ~
    \hat{y}_{i_1} = \max(y_1\times (1-r_i),~0) \\
    \hat{x}_{i_2} = \min(x_2\times (1+r_i),~\hat{W}),& ~\hat{y}_{i_2} = \min(y_2\times (1+r_i),~\hat{H}) \\
    r_i = r_{i-1}+\frac{\rho}{K},&~ r_0=0,
\end{split}
\end{align}
where $r$ represents expansion ratio, $\hat{W}$ and $\hat{H}$ represent input image width and height. We find that different hyperparameter K values only caused minor differences in performance across three datasets. For consistency, we use $\rho=0.4, K=5$ in this paper. In this scenario, the RPR module brings negligible extra memory consumption during inference (about 0.01 seconds per iteration per device). Next, ROIAlign \cite{he2017mask} extracts $\text{ROI}^{\text{rec}}_i$ from feature map $F$ based on each expanded proposal box $B^{\text{expand}}_i$ :
\begin{align}
    \text{ROI}^{\text{rec}}_i = \text{ROIAlign}(F,B^{\text{expand}}_i),~ i=1,\dots,K.
\end{align}

\subsubsection{Attentive FFN} The ROI features  ($\text{ROIs}^{\text{rec}}$) extracted from progressive ROIAlign module are then fed into the proposed attentive feed-forward network (FFN) for region proposal box rectification. As shown in Fig.~\ref{Figure2}c, the attentive FFN module employs a self-attention mechanism \cite{vaswani2017attention} to help build pixel relationships in spatial space of $\text{ROIs}^{\text{rec}}$. 

Before self-attention module, we separate the ROIs$^{\text{rec}}$ 
into two sets: ROI$^{\text{ori}}$ and ROIs$^{\text{expand}}$. ROI$^{\text{ori}}\in\mathbb{R}^{N\times C\times D}$ where $D=14\times 14$ represents the ROI feature extracted using original region proposal box. ROIs$^{\text{expand}}\in\mathbb{R}^{N\times C\times (K-1)D}$ involves the ROI features from expanded proposal boxes, note that we stack the ROIs in the spatial dimension. We use ROI$^{\text{ori}}$ as query features, and  ROIs$^{\text{expand}}$ as value and key features. To perform self-attention, we first project channel dimension $C=256$ of query, key, value features to $\hat{C}=64$ with $1\times1$ convolutional layers. We use $\mathcal{Q}\in \mathbb{R}^{N\times \hat{C}\times D}, \mathcal{K}\in \mathbb{R}^{N\times \hat{C}\times (K-1)D}$ and $\mathcal{V}\in \mathbb{R}^{N\times \hat{C}\times (K-1)D}$ to represent the projected query, key and value features. The attentive ROI features ($\text{ROI}^{att}\in \mathbb{R}^{N\times \hat{C}\times D}$) from the self-attention module is then obtained as:
\begin{gather}
\text{ROI}^{att} = \mathcal{Q}+\mathcal{V}\cdot\text{softmax}(\frac{(\mathcal{Q}^T\mathcal{K})^T}{\sqrt{\hat{C}}}).
\end{gather}

Finally, convolutional layers and fully connected (FC) layers are utilized to linearly transform the $\text{ROI}^{att}$ to rectified proposal boxes $\mathcal{B}^{\text{rec}}\in \mathbb{R}^{N\times 4}$. We use smooth $L_1$ loss \cite{girshick2015fast} for box regression, we term the loss as $L_{\text{RPR}}$.

\section{Experiment}
\subsection{Datasets}
We evaluate the performance of proposed method on three datasets: urothelial cell, plant phenotyping and DSB2018. Plant phenotyping and DSB2018 are public datasets. For each dataset, we use 70\%, 15\%, 15\% of images for training, validation, and testing.
(1) \textbf{Urothelial Cell}. Instance segmentation of urothelial cells is of great importance in urine cytology for urothelial carcinoma detection. The urothelial cell dataset contains 336 pathological images with a resolution of $1024\times 1024$. The images are cropped from ThinPrep slide images and are annotated by three experts. 
(2) \textbf{Plant Phenotyping}. The plant phenotyping dataset \cite{minervini2016finely} contains 535 top-down view plant leave images with various resolutions. 
(3) \textbf{DSB2018}. The Data Science Bowl 2018 (DSB2018\footnote{https://www.kaggle.com/c/data-science-bowl-2018}) dataset consists of 670 annotated cell nuclei images with different sizes.  

\begin{table*}[h]
\centering
\caption{Quantitative evaluation results of baseline methods with and without proposed region proposal rectification (RPR) module on three biological datasets.}
\label{Table1}
\resizebox{\textwidth}{!}{
\begin{tabular}{l|c|ccc|ccc}
\hline
\multirow{2}{*}{Methods} & \multirow{2}{*}{Datasets} & \multicolumn{3}{c|}{${\text{AP}^\mathrm{bbox}}$ ($\%$)} & \multicolumn{3}{c}{${\text{AP}^\mathrm{mask}}$ ($\%$)}\\
\cline{3-8}   & & ${\text{AP}^\mathrm{bbox}}$ & ${\text{AP}_\mathrm{0.5}^\mathrm{bbox}}$ & ${\text{AP}_\mathrm{0.75}^\mathrm{bbox}}$ & ${\text{AP}^\mathrm{mask}}$ & ${\text{AP}_\mathrm{0.5}^\mathrm{mask}}$ & ${\text{AP}_\mathrm{0.75}^\mathrm{mask}}$ \\ 
\hline
Mask R-CNN   ~\cite{he2017mask}        & \multirow{4}{*}{Plant}    & 58.10  & 88.14  & 65.85  & 56.15  & 84.39  & 65.29  \\
\textbf{Mask R-CNN+RPR   }                                           & & \textbf{60.69}  & \textbf{88.23}  & \textbf{67.32}  & \textbf{57.56}  & \textbf{84.49}  & \textbf{65.93}  \\
CenterMask  ~\cite{lee2020centermask}                           & & 55.15  & 87.76  & 60.33  & 50.39  & \textbf{83.43}  & 56.45  \\
\textbf{CenterMask+RPR  }                                           & & \textbf{59.78}  & \textbf{87.79}  & \textbf{66.94}  & \textbf{51.52}  & 83.15  & \textbf{58.44}  \\
\hline
Mask R-CNN   ~\cite{he2017mask}        & \multirow{4}{*}{DSB2018}  & 53.36  & 78.82  & 60.84  & 50.46  & \textbf{77.63}  & 57.08  \\ 
\textbf{Mask R-CNN+RPR }                                             & & \textbf{54.02}  & \textbf{78.88}  & \textbf{61.10}  & \textbf{50.75}  & 77.62  & \textbf{57.25} \\ 
CenterMask  ~\cite{lee2020centermask}                           & & 55.30  & 79.71  & 62.82  & 51.84  & \textbf{78.51}  & 59.62  \\
\textbf{CenterMask+RPR  }                                           & & \textbf{55.82}  & \textbf{79.75}  & \textbf{63.56}  & \textbf{52.09}  & 78.46  & \textbf{59.63}  \\ 
\hline
Mask R-CNN   ~\cite{he2017mask}        & \multirow{4}{*}{Urothelial Cell}& 76.56  & 91.16  & 87.65  & 77.35  & 91.16  & \textbf{89.02}  \\ 
\textbf{Mask R-CNN+RPR}                                              & & \textbf{77.98}  & \textbf{91.16}  & \textbf{88.49}  & \textbf{77.65}  & \textbf{91.16}  & 88.99  \\  
CenterMask  ~\cite{lee2020centermask}                           & & 76.11  & 94.52  & 88.03  & 76.87  & 94.55  & 88.51  \\
\textbf{CenterMask+RPR }                                            & & \textbf{80.76}  & \textbf{94.66}  & \textbf{91.00}  & \textbf{78.81}  & \textbf{94.66}  & \textbf{91.05}  \\
\hline
\end{tabular}} 
\end{table*}

\subsection{Experimental Details}
We follow the same training skills and hyper-parameter settings as Mask R-CNN \cite{he2017mask} and CenterMask \cite{lee2020centermask}. In particular, we use stochastic gradient descent (SGD) algorithm as optimizer with an initial learning rate of 0.001. We use random flipping as data augmentation and train the network for 10k iterations with a batch size of 16. The weights of the backbone network are pre-trained on ImageNet dataset. We use 8 Quadro RTX 8000 GPUs for training. The overall object loss function is $L = L_\text{box} + L_\text{mask} + L_\text{RPR}$, where $L_\text{box}$ and $L_\text{mask}$ represent the proposal box loss and mask prediction loss. We use the same $L_\text{box}$ and $L_\text{mask}$ losses as Mask R-CNN \cite{he2017mask} and CenterMask \cite{lee2020centermask}.

\subsection{Evaluation Metric} 
We report the average precision (AP) \cite{he2017mask} averaged over IoU threshold ranges from 0.5 to 0.95 at an interval of 0.05 as evaluation metric:
\begin{equation}
    AP=\frac{1}{10}\sum_{t=0.5:0.05:0.95}AP_t,
\end{equation}
where $t$ indicates the IoU threshold between predicted bbox/mask and ground truth bbox/mask. We use $\text{AP}^{\text{bbox}}$ and $\text{AP}^{\text{mask}}$ to represent the box- and mask-level AP. We additionally report AP at IoU threshold of 0.5 and 0.75 in Table~\ref{Table1}.

\begin{figure}[thb!]
\begin{center}
\includegraphics[width=0.8\linewidth]{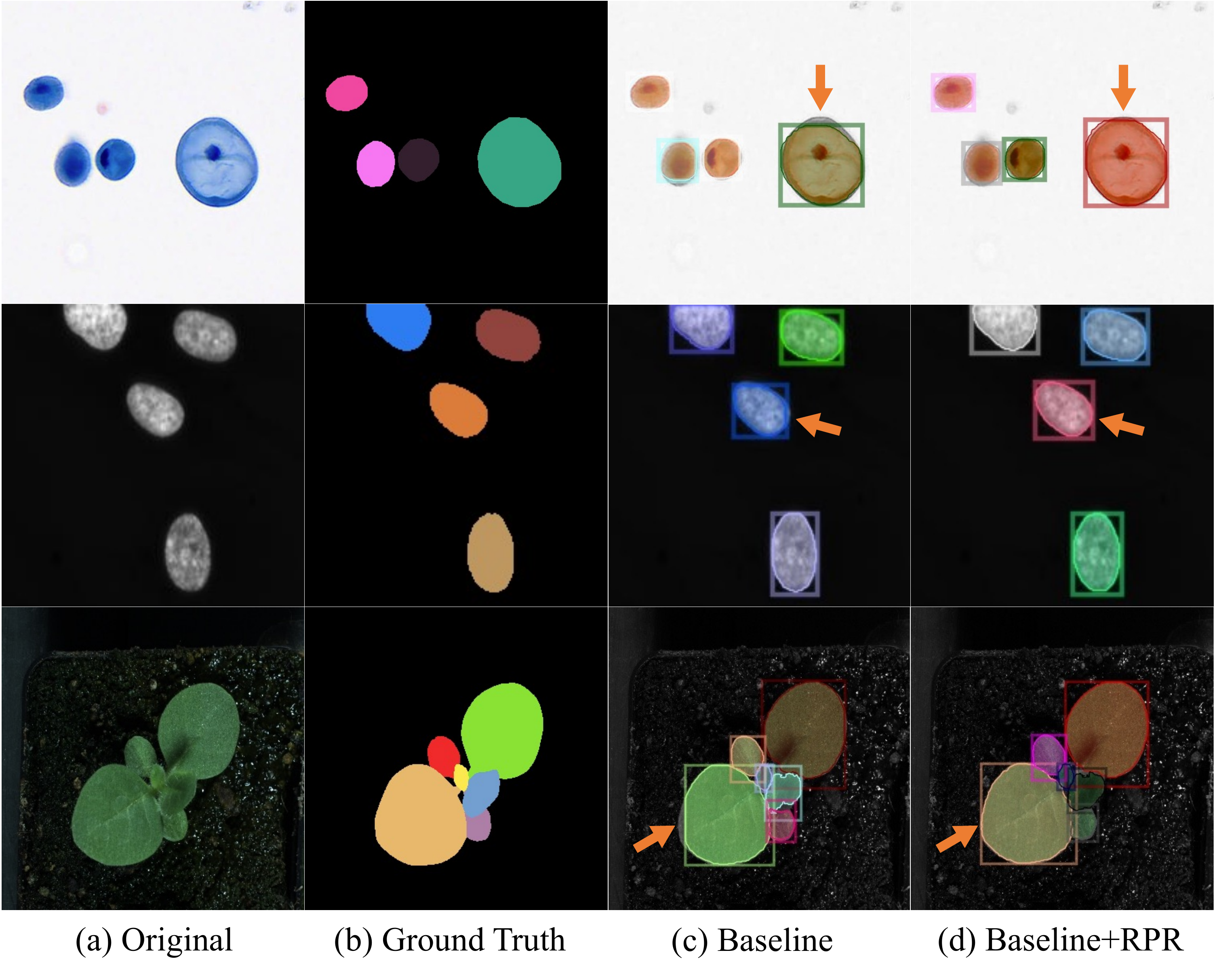}
\end{center}
   \caption{The qualitative results of baseline methods (i.e., Mask R-CNN \cite{he2017mask} and CenterMask \cite{lee2020centermask}) with and without region proposal rectification (RPR) module. From left to right, we show the original image, the ground-truth annotation, results by either Mask R-CNN or CenterMask without RPR, and results with RPR. From top to bottom, we illustrate the images from urothelial cell, DSB2018, and plant datasets respectively. 
   }
\label{Figure3}
\end{figure}

\section{Results and Discussion} 
The qualitative and quantitative instance segmentation results are shown in Fig.~\ref{Figure3} and Table~\ref{Table1} respectively. In particular, we have compared the results of baseline methods (i.e., Mask R-CNN \cite{he2017mask} and CenterMask \cite{lee2020centermask}) with and without the proposed region proposal rectification (RPR) on Urothelial Cell, Plant Phenotyping and DSB2018 datasets.

As shown in Table~\ref{Table1}, with region proposal rectification, the performances of object detection ($\text{AP}^{\text{bbox}}$) and instance segmentation ($\text{AP}^{\text{mask}}$) are consistently improved for baseline methods on three biological datasets. Specifically, Mask R-CNN with RPR improves ${\text{AP}^\mathrm{bbox}}$ by 2.59, 0.66, 1.42 points and ${\text{AP}^\mathrm{mask}}$ by 1.41, 0.29, 0.3 points on Plant, DSB2018, and Urothelial Cell datasets, respectively. CenterMask with RPR improves ${\text{AP}^\mathrm{bbox}}$ by 4.63, 0.52, 4.65 points and ${\text{AP}^\mathrm{mask}}$ by 1.13, 0.25, 1.94 points on the three biological datasets. From qualitative results in Fig.~\ref{Figure3}, we observe that the incomplete segmentation mask problem is remarkably relieved with the proposed region proposal rectification.

\begin{figure}[h]
\begin{center}
\includegraphics[width=0.47\linewidth]{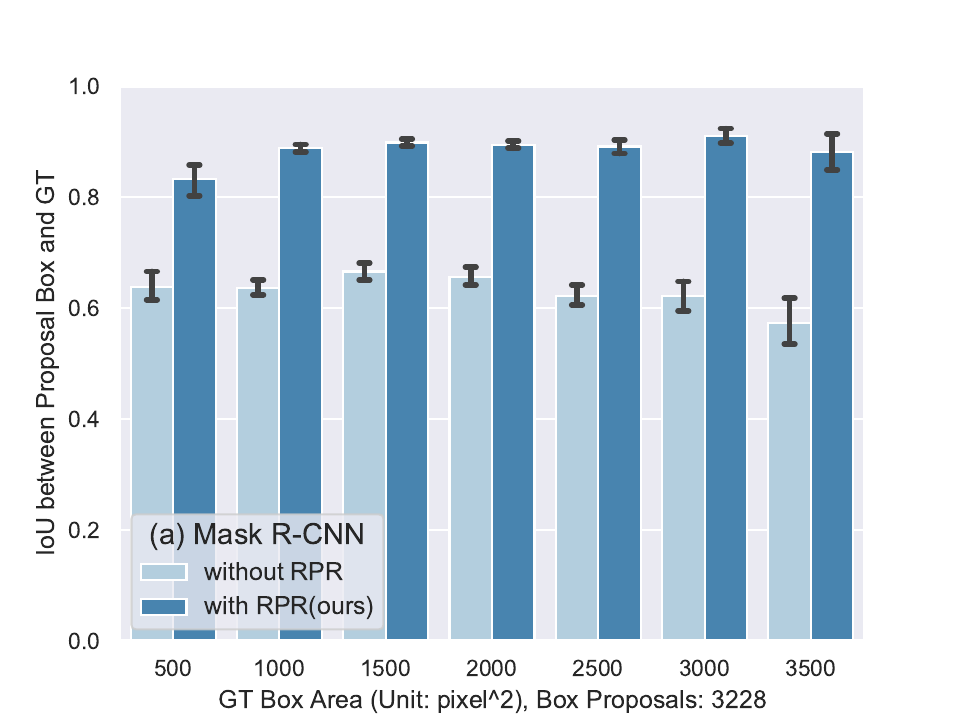}\hspace{-2mm}
\includegraphics[width=0.47\linewidth]{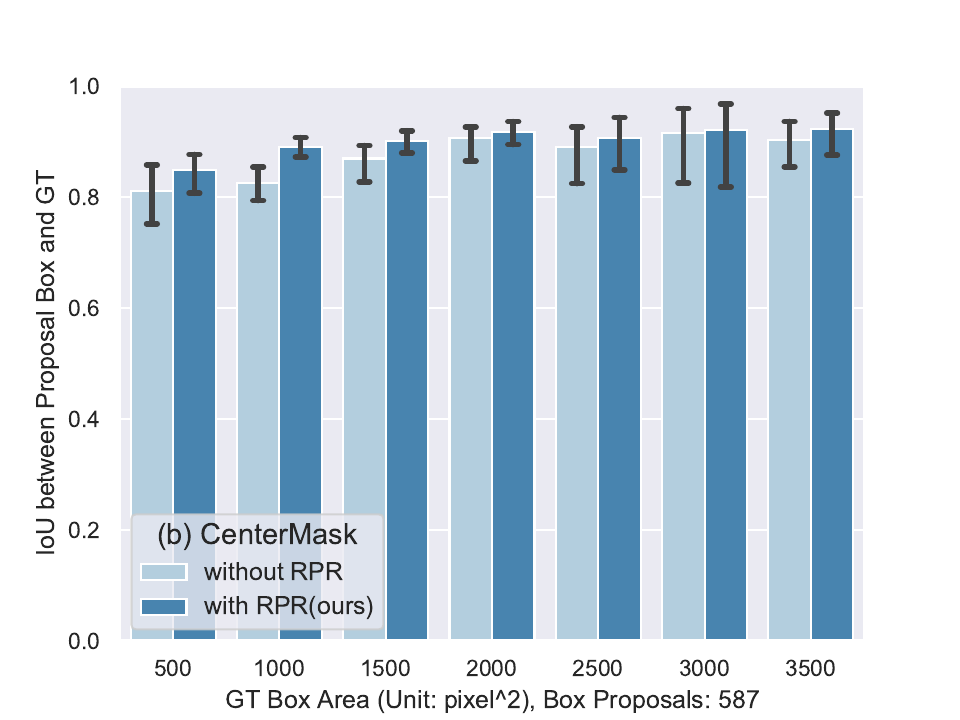}\hspace{-2mm}
   \caption{Histogram of intersection-over-union (IoU) between region proposal boxes and ground truth (GT) boxes along with GT box areas on urothelial cell testing dataset. The box area unit is pixel$^2$. Error bars (mean $\pm$ std) are added.}
\label{Figure4}
\end{center}
\end{figure}

To investigate the effect of region proposal rectification (RPR) module, we exhibit the histogram of IoU between region proposal boxes and ground truth boxes on urothelial cell testing datasets in Fig.~\ref{Figure4}. In particular, we split the ground truth boxes into 7 bins with box areas range from 500 pixels$^2$ to 3500 pixels$^2$ at an interval of 500 pixels$^2$. For each bin, we calculate the IoUs between the ground truth boxes and corresponding proposal boxes. We add error bars (means $\pm$ std) on histograms. As shown in Fig.~\ref{Figure4}a, Mask R-CNN with RPR shows significant improvement over Mask R-CNN without RPR in box IoU, suggesting that the RPR module corrects region proposal boxes significantly.  

The results demonstrate that introduction of neighbor spatial knowledge to ROI features is effective for box rectification. Note that this improvement does not relate with the AP$^{\text{bbox}}$ in Table~\ref{Table1} as we do not employ non-maximum-suppression (NMS) here since it helps us better observe the rectification results of the RPR module. In Fig.~\ref{Figure4}b, CenterMask with RPR also shows superiority in box IoU especially on cells with smaller box areas. The improvement is not as significant as Mask R-CNN, one possible reason would be that Mask R-CNN uses anchor-based object detector and provides a large number of proposal boxes (e.g., 3228 boxes). On the contrary, CenterMask employs anchor-free FCOS to provide a small number of precise proposal boxes (e.g., 587 boxes). Before NMS operation, Mask-RCNN's proposal locations are not as precise as CenterMask. However, with RPR, the gaps become small.

\begin{figure} [h]
\begin{center}
\includegraphics[width=0.85\linewidth]{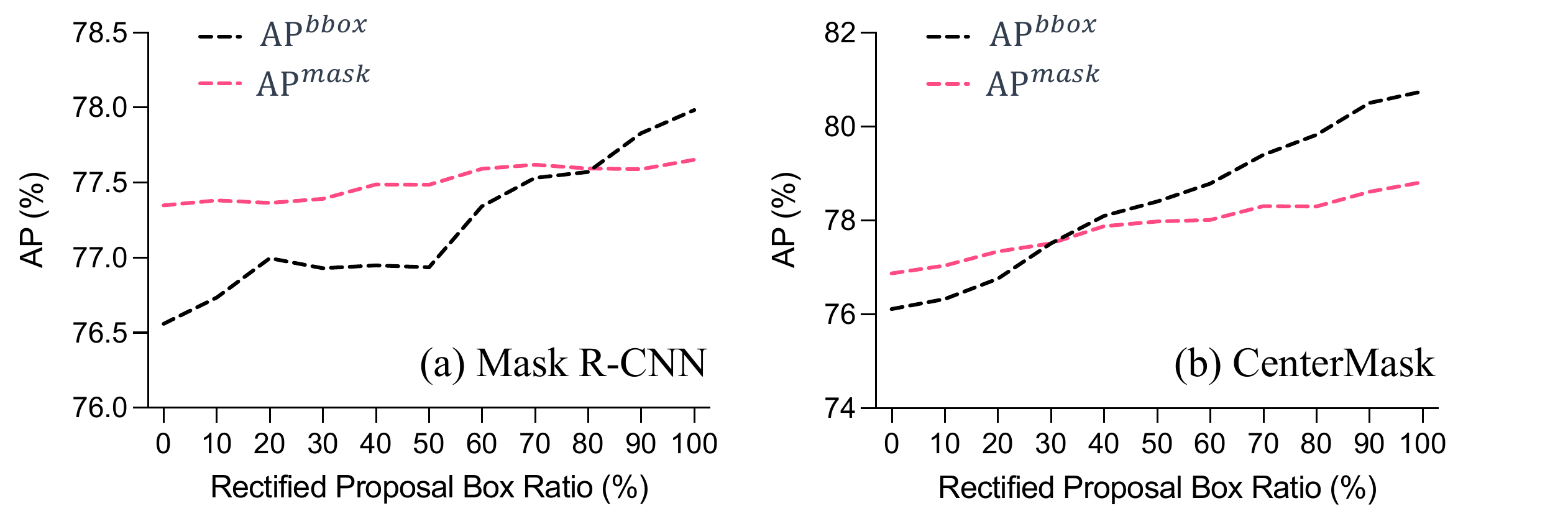}
\end{center}
  \caption{${\text{AP}^\mathrm{bbox}}$ and ${\text{AP}^\mathrm{mask}}$ along with various fractions of rectified proposal boxes. 
  }
\label{Figure5}
\end{figure} 

Considering the improvement brought by our RPR module is mainly around the object’s boundary, therefore, we present AP$^{\text{bbox}}$ and AP$^{\text{mask}}$ along with various fractions of rectified region proposal boxes in Fig.~\ref{Figure5}. As can be seen, both AP$^{\text{bbox}}$  and AP$^{\text{mask}}$ curves increase with fractions of rectified proposal boxes in the baseline methods. The improvement is more significant in AP$^{\text{bbox}}$ compared to AP$^{\text{mask}}$. Fig.~\ref{Figure5} further explains the effectiveness of the proposed RPR module. Addressing the robustness of the RPR module in extreme cases with blurred cell boundaries can be a promising direction for future research.

\section{Conclusion}
In this study, we proposed a region proposal rectification (RPR) module that addresses the challenging incomplete segmentation problem in the top-down instance segmentation of biological images. The proposed approach offers new perspectives to redefine the procedure of biological instance mask generation. 
The RPR module shows significant improvement in both anchor-based and anchor-free top-down instance segmentation baselines. The effective results on three biological datasets demonstrate that the proposed approach can be served as a solid baseline in instance segmentation of biological images.

\bibliographystyle{splncs04}
\typeout{}
\bibliography{mybib}

\end{document}